\documentclass[11pt,a4paper]{article}
\usepackage[hyperref]{emnlp2020}
\usepackage{times}
\usepackage{latexsym}

\usepackage{multirow}
\usepackage{booktabs}
\usepackage{algorithmic}
\usepackage{amsmath}
\usepackage{esvect}
\usepackage{latexsym}
\usepackage{CJK}
\usepackage{enumitem}
\usepackage{mathrsfs}
\usepackage{amsfonts}
\usepackage{bm}
\usepackage{colortbl}
\usepackage{graphicx}
\usepackage{subfigure}
\usepackage{soul}
\usepackage{url}
\usepackage{amsmath}
\usepackage[utf8]{inputenc}
\usepackage{xcolor}
\usepackage{algorithm}
\usepackage{amssymb}
\usepackage{makecell}
\usepackage{xspace}
\usepackage{array}
\usepackage[normalem]{ulem}
\usepackage{footmisc}
\usepackage{colortbl}
\usepackage{epstopdf}
\usepackage{setspace}
\usepackage{pifont}

\newcommand{\ie}{\emph{i.e.,}\xspace}

\usepackage{microtype}

\makeatletter 
\newcommand\figcaption{\def\@captype{figure}\caption} 
\newcommand\tabcaption{\def\@captype{table}\caption} 

\title{VMSMO: Learning to Generate Multimodal Summary for \\ Video-based News Articles}
\aclfinalcopy
\author{
Mingzhe Li$^{1,2,}$\thanks{\;\;Equal contribution. Ordering is decided by a coin flip.}$^{\;}$,
Xiuying Chen$^{1,2,}$\footnotemark[1],
Shen Gao$^{2}$, 
\\ \bf Zhangming Chan$^{1,2}$, 
Dongyan Zhao$^{1,2}$ \and 
Rui Yan$^{1,2,}$\thanks{\;\;Corresponding author.} \\
$^1$ Center for Data Science, AAIS, Peking University,Beijing,China \\
$^2$ Wangxuan Institute of Computer Technology, Peking University,Beijing,China \\
{\tt \{li\_mingzhe,xy-chen,shengao,zhaody,ruiyan\}@pku.edu.cn} \\
}
\date{}

\begin{document}
\maketitle
\begin{abstract}
A popular multimedia news format nowadays is providing users with a lively video and a corresponding news article, which is employed by influential news media including CNN, BBC, and social media including Twitter and Weibo.
In such a case, automatically choosing a proper cover frame of the video and generating an appropriate textual summary of the article can help editors save time, and readers make the decision more effectively.
Hence, in this paper, we propose the task of Video-based Multimodal Summarization with Multimodal Output (VMSMO) to tackle such a problem.
The main challenge in this task is to jointly model the temporal dependency of video with semantic meaning of article.
To this end, we propose a Dual-Interaction-based Multimodal Summarizer (DIMS), consisting of a dual interaction module and multimodal generator.
In the dual interaction module, we propose a conditional self-attention mechanism that captures local semantic information within video and a global-attention mechanism that handles the semantic relationship between news text and video from a high level.
Extensive experiments conducted on a large-scale real-world VMSMO dataset\footnote{https://github.com/yingtaomj/VMSMO} show that DIMS achieves the state-of-the-art performance in terms of both automatic metrics and human evaluations.

\end{abstract}

\begin{figure}
    \centering
    \includegraphics[scale=0.7]{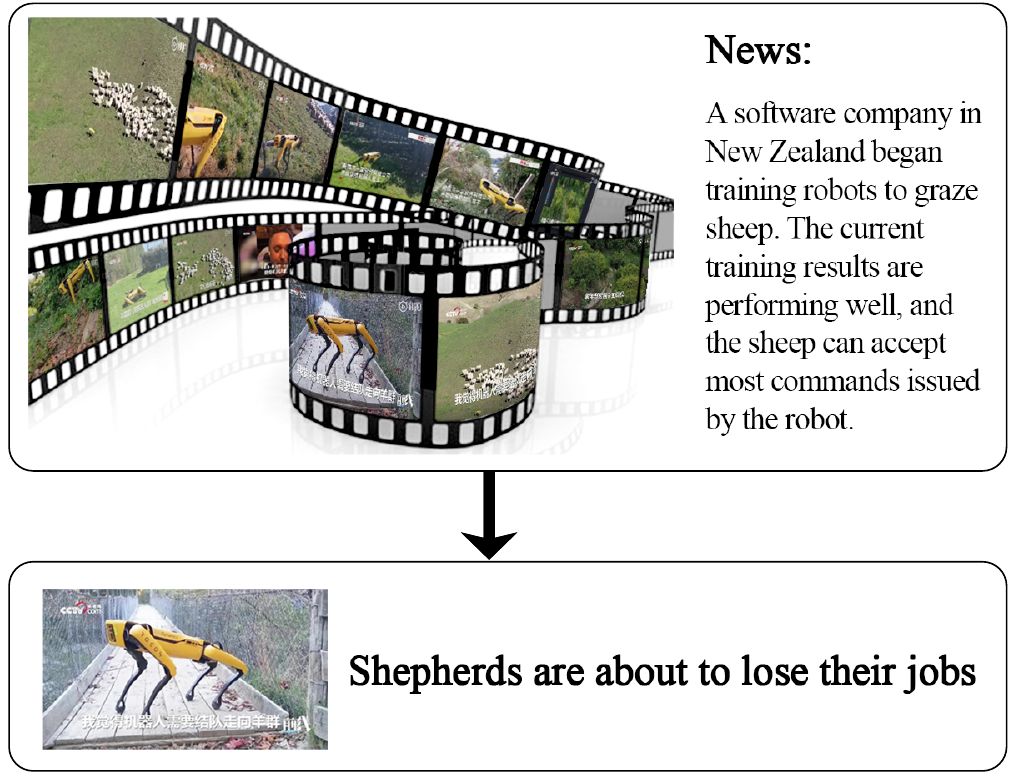}
    \caption{
        An example of video-based multimodal summarization with multimodal output.
    }
    \label{fig:intro}
    \vspace{-5mm}
\end{figure} 

\section{Introduction}
Existing experiments \cite{li2017multi} have proven that multimodal news can significantly improve users’ sense of satisfaction for informativeness.
As one of these multimedia data forms, introducing news events with video and textual descriptions is becoming increasingly popular, and has been employed as the main form of news reporting by news media including BBC, Weibo, CNN, and Daily Mail.
An illustration is shown in Figure~\ref{fig:intro}, where the news contains a video with a cover picture and a full news article with a short textual summary.
In such a case, automatically generating multimodal summaries, \ie choosing a proper cover frame of the video and generating an appropriate textual summary of the article can help editors save time and readers make decisions more effectively.

There are several works focusing on multimodal summarization.
The most related work to ours is \cite{zhu2018msmo}, where they propose the task of generating textual summary and picking the most representative picture from 6 input candidates.
However, in real-world applications, the input is usually a video consisting of hundreds of frames.
Consequently, the temporal dependency in a video cannot be simply modeled by static encoding methods.
Hence, in this work, we propose a novel task, Video-based Multimodal Summarization with Multimodal Output (VMSMO), which selects cover frame from news video and generates textual summary of the news article in the meantime.
    

The cover image of the video should be the salient point of the whole video, while the textual summary should also extract the important information from source articles.
Since the video and the article focus on the same event with the same report content, these two information formats complement each other in the summarizing process.
However, how to fully explore the relationship between temporal dependency of frames in video and semantic meaning of article still remains a problem, since the video and the article come from two different space.

Hence, in this paper, we propose a model named \emph{Dual-Interaction-based Multimodal Summarizer (DIMS)}, which learns to summarize article and video simultaneously by conducting a dual interaction strategy in the process.
Specifically, we first employ Recurrent Neural Networks (RNN) to encode text and video.
Note that by the encoding RNN, the spatial and temporal dependencies between images in the video are captured. 
Next, we design a dual interaction module to let the video and text fully interact with each other. 
Specifically, we propose a conditional self-attention mechanism which learns local video representation under the guidance of article, and a global-attention mechanism to learn high-level representation of video-aware article and article-aware video.
Last, the multimodal generator generates the textual summary and extracts the cover image based on the fusion representation from the last step.
To evaluate the performance of our model, we collect the first large-scale news article-summary dataset associated with video-cover from social media websites.
Extensive experiments on this dataset show that DIMS significantly outperforms the state-of-the-art baseline methods in commonly-used metrics by a large margin.

To summarize, our contributions are threefold: 

$\bullet$ We propose a novel Video-based Multimodal Summarization with Multimodal Output (VMSMO) task which chooses a proper cover frame for the video and generates an appropriate textual summary of the article.

$\bullet$ We propose a Dual-Interaction-based Multimodal Summarizer (DIMS) model, which jointly models the temporal dependency of video with semantic meaning of article, and generates textual summary with video cover simultaneously.

$\bullet$ We construct a large-scale dataset for VMSMO, and experimental results demonstrate that our model outperforms other baselines in terms of both automatic and human evaluations.

\begin{figure*}
    \centering
    \includegraphics[scale=0.52]{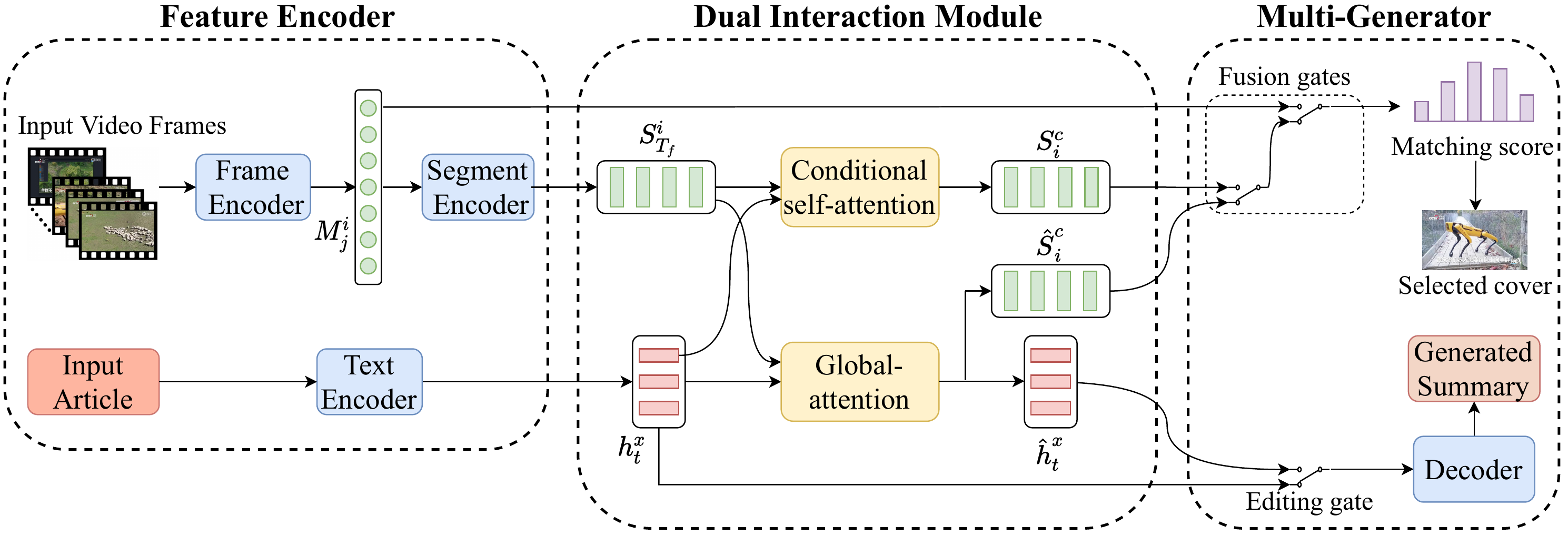}
    \caption{
        Overview of DIMS. We divide our model into three parts: (1) \textit{Feature Encoder} encodes the input article and video separately; (2) \textit{Dual Interaction Module} learns fused representation of video and article from different level; (3) \textit{Multi-Generator} generates the textual summary and chooses the video cover simultaneously.
    }
    \label{fig:overview}
\end{figure*}

\section{Related Work}
Our research builds on previous works in three fields: text summarization, multimodal summarization, and visual question answering.

\noindent \textbf{Text Summarization.}
Our proposed task bases on text summarization, the methods of which can be divided into extractive and abstractive methods~\cite{Gao2020From}.
Extractive models~\cite{zhang2018neural, Narayan2018RankingSF,chen2018iterative,Luo2019ReadingLH,Xiao2019ExtractiveSO} directly pick sentences from article and regard the aggregate of them as the summary.
In contrast, abstractive models ~\cite{sutskever2014sequence,See2017GetTT, Wenbo2019ConceptPN, Gui2019AttentionOF,gao2019write,chen2019learning,gao2019abstractive} generate a summary from scratch and the abstractive summaries are typically less redundant.


\noindent \textbf{Multimodal Summarization.}
A series of works \cite{li2017multi,li2018multi,palaskar2019multimodal,chan2019stick,chen2019rpm,gao2020learning} focused on generating better textual summaries with the help of multimodal input.
Multimodal summarization with multimodal output is relatively less explored.
\citet{zhu2018msmo} proposed to jointly generate textual summary and select the most relevant image from 6 candidates. 
Following their work, \citet{zhu3multimodal} added a multimodal objective function to use the loss from the textual summary generation and the image selection.
However, in the real-world application, we usually need to choose the cover figure for a continuous video consisting of hundreds of frames.
Consequently, the temporal dependency between frames in a video cannot be simply modeled by several static encoding methods.


\noindent \textbf{Visual Question Answering.}
Visual Question Answering (VQA) task is similar to our task in taking images and a corresponding text as input.
Most works consider VQA task as a classification problem and the understanding of image sub-regions or image recognition becomes particularly important~\cite{goyal2017making, malinowski2015ask, wu2016ask, xiong2016dynamic}.
As for the interaction models, one of the state-of-the-art VQA models \cite{li2019beyond} proposed a positional self-attention with a co-attention mechanism, which is faster than the recurrent neural network (RNN). 
\citet{guo2019image} devised an image-question-answer synergistic network, where candidate answers are coarsely scored according to their relevance to the image and question pair and answers with a high probability of being correct are re-ranked by synergizing with image and question.

\section{Problem Formulation}
Before presenting our approach for the VMSMO, we first introduce the notations and key concepts.
For an input news article $X = \{x_1, x_2, \dots, x_{T_d}\}$  which has ${T_d}$ words, we assume there is a ground truth textual summary $Y = \{y_1, y_2, \dots, y_{T_y}\}$  which has ${T_y}$ words. 
Meanwhile, there is a news video $V$ corresponding to the article, and we assume there is a ground truth cover picture $C$ that extracts the most important frame from the video content. 
For a given article $X$ and the corresponding video $V$, our model emphasizes salient parts of both inputs by conducting deep interaction.
The goal is to generate a textual summary $Y^{'}$ that successfully grasp the main points of the article and choose a frame picture $C^{'}$ that covers the gist of the video.

\section{Model}
\subsection{Overview}
In this section, we propose our Dual Interaction-based Multimodal Summarizer (DIMS), which can be divided into three parts in Figure~\ref{fig:overview}:

$\bullet$ \textit{Feature Encoder} is composed of a text encoder and a video encoder which encodes the input article and video separately.

$\bullet$ \textit{Dual Interaction Module} conducts deep interaction, including conditional self-attention and global-attention mechanism between video segment and article to learn different levels of representation of the two inputs.

$\bullet$ \textit{Multi-Generator} generates the textual summary and chooses the video cover by incorporating the fused information.

\subsection{Feature Encoder}
\subsubsection{Text encoder}
To model the semantic meaning of the input news text $X = \{x_1, x_2, \dots, x_{T_d}\}$, we first use a word embedding matrix $e$ to map a one-hot representation of each word $x_{i}$ into to a high-dimensional vector space.
Then, in order to encode contextual information from these embedding representation, we use bi-directional recurrent neural networks (Bi-RNN) \cite{Hochreiter1997LongSM} to model the temporal interactions between words:
\begin{align}
h^x_t &= \text{Bi-RNN}_\text{X}(e(x_t), h^x_{t-1}),
\end{align}
where $h^x_t$ denotes the hidden state of $t$-th step in Bi-RNN for $X$. 
Following \cite{See2017GetTT, ma2018hierarchical}, we choose the long short-term memory (LSTM) as the Bi-RNN cell.

\subsubsection{Video Encoder}
A news video usually lasts several minutes and consists of hundreds of frames.
Intuitively, a video can be divided into several segments, each of which corresponds to different content.
Hence, we choose to encode video hierarchically.
More specifically, we equally divide frames in the video into several segments and employ a low-level frame encoder and a high-level segment encoder to learn hierarchical representation.

\textbf{Frame encoder.}
We utilize the Resnet-v1 model \cite{he2016deep} to encode frames to alleviate gradient vanishing \cite{he2016deep} and reduce computational costs:
\begin{align}
    O_j^i&=\text{Resnet-v1}(m_j^i),\\
    M_j^i&=\operatorname{relu}\left(F_v(O_j^i)\right),
\end{align}
where $m_j^i$ is the $j$-th frame in $i$-th segment and $F_{v}(\cdot)$ is a linear transformation function.

\textbf{Segment encoder.}
As mentioned before, it is important to model the continuity of images in video, which cannot be captured by a static encoding strategy. 
We employ RNN network as segment encoder due to its superiority in exploiting the temporal dependency among frames \citet{zhao2017hierarchical}:
\begin{align}
    S_j^i = \text{Bi-RNN}_\text{S}(M_j^i, S^i_{j-1}).
\end{align}
$S_j^i$ denotes the hidden state of $j$-th step in Bi-RNN for segment $s_i$, and the final hidden state $S_{T_f}^i$ denotes the overall representation of the segment $s_i$, where $T_f$ is the number of frames in a segment.

\begin{figure}
    \centering
    \includegraphics[scale=0.43]{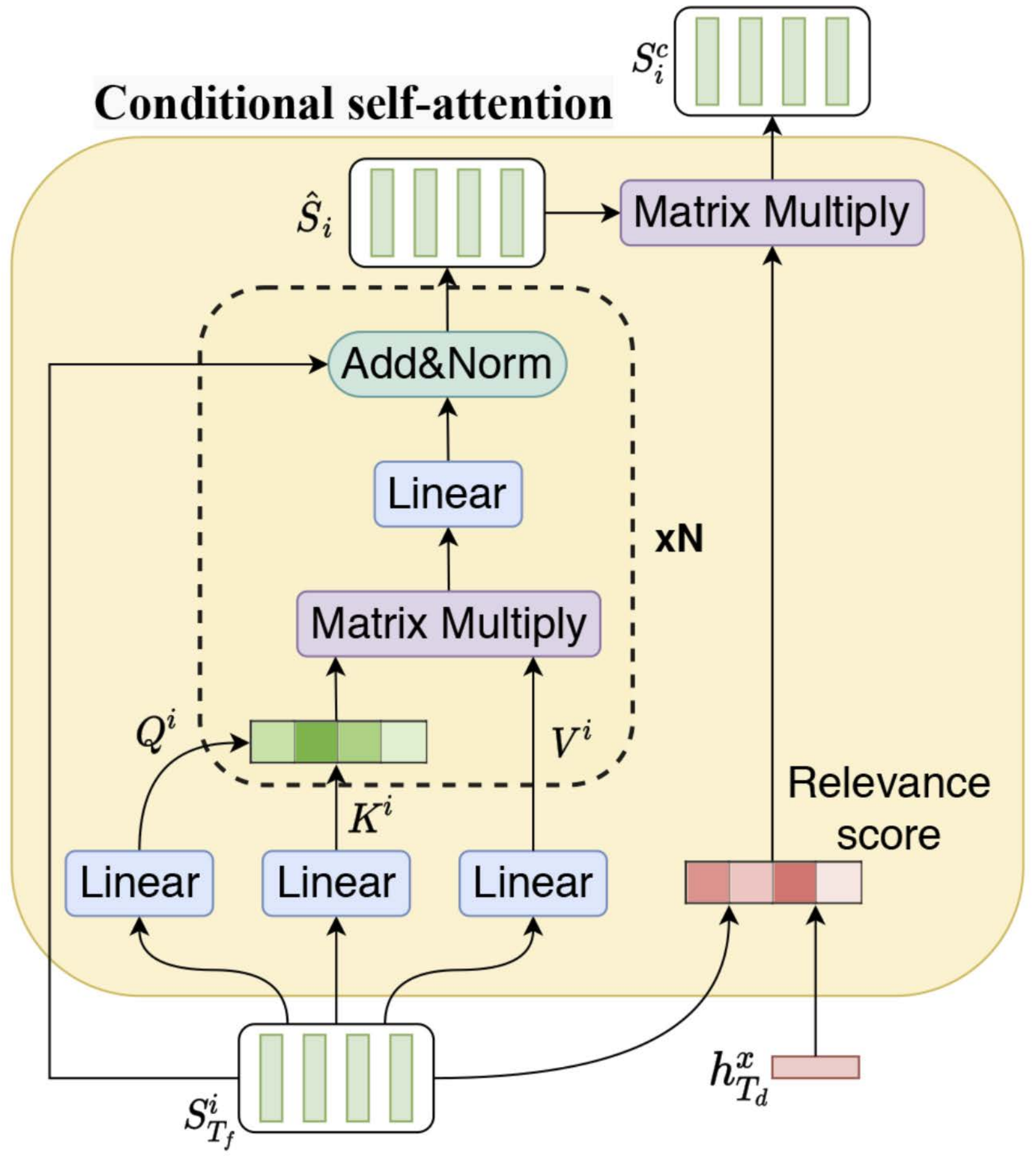}
    \caption{
        Conditional self-attention module, which captures local semantic information within video segments under the guidance of article representation.
    }
    \label{fig:cond}
\end{figure}

\subsection{Dual Interaction Module}
The cover image of the video should contain the key point of the whole video, while the textural summary should also cover extract the important information from source articles.
Hence, these two information formats complement each other in the summarizing process.
In this section, we conduct a deep interaction between the video and article to jointly model the temporal dependency of video and semantic meaning of text.  
The module consists of a conditional self-attention mechanism that captures local semantic information within video segments and a global-attention mechanism that handles the semantic relationship between news text and video from a high level.

\textbf{Conditional self-attention mechanism.}
Traditional self-attention can be used to obtain contextual video representation due to its flexibility in relating two elements in a distance-agnostic manner.
However, as illustrated in \citet{xie2020conditional}, the semantic understanding often relies on more complicated dependencies than the pairwise one, especially conditional dependency upon a given premise. 
Hence, in the VMSMO task, we capture the local semantic information of video conditioned on the input text information.

Our conditional self-attention module shown in Figure~\ref{fig:cond} is composed of a stack of N identical layers and a conditional layer. 
The identical layer learns to encode local video segments while the conditional layer learns to assign high weights to the video segments conditioned on their relationship to the article.
We first use a fully-connected layer to project each segment representation $S_{T_f}^i$ into the query $Q^i$, key $K^i$, and value $V^i$.
Then, the scaled dot-product self-attention is defined as:
\begin{align}
\alpha_{i, j} &=\frac{\exp \left(Q^{i} K^{j}\right)}{\sum\nolimits_{n=1}^{T_{s}} \exp \left(Q^{i} K^{n}\right)}, \\
\hat{S}_{i} &=\sum\nolimits_{j=1}^{T_{s}} \frac{\alpha_{i, j} V^{j}}{\sqrt d},
\end{align}
where $d$ stands for hidden dimension and $T_{s}$ is the segment number in a video.
$\hat{S}_{i}$ is then fed into the feed-forward sub-layer including a residual connection~\cite{he2016deep} and layer normalization~\cite{ba2016layer}.

Next, we highlight the salient part of the video under the guidance of article. 
Taking the article information $h_{T_d}^x$ as condition, the attention score on each original segment representation $S_{T_f}^i$ is calculated as:
\begin{align}
    \beta_{i} &=\sigma\left(F_s(S_{T_f}^i h_{T_d}^x)\right).
\end{align}
The final conditional segment representation $S^{c}_{i}$ is denoted as $\beta_{i} \hat{S}_{i}$.

\textbf{Global-attention mechanism.}
The global-attention module grounds the article representation on the video segments and fuses the information of the article into the video, which results in an article-aware video representation and a video-aware article representation.
Formally, we utilize a two-way attention mechanism to obtain the co-attention between the encoded text representation $h^x_t$ and the encoded segment representation $S_{T_f}^i$:
\begin{align}
\label{global}
    E^{t}_{i} &= F_{h}(h^x_t) \left(F_{t}(S_{T_f}^i)\right)^T.
\end{align}
We use $E^{t}_{i}$ to denote the attention weight on the $t$-th word by the $i$-th video segment.
To learn the alignments between text and segment information, the global representations of video-aware article $\hat{h}^x_t$ and article-aware video $\hat{S}_i^c$ are computed as:
\begin{align}
    \hat{h}^x_t &= \sum\nolimits_{i=1}^{T_d} E^{t}_{i} S_{T_f}^i,\\
    \hat{S}_i^c &= \sum\nolimits_{t=1}^{T_s} \left( {E^{t}_{i}}\right )^T h^x_t.
\end{align}


\subsection{Multi-Generator}
In the VMSMO task, the multi-generator module not only needs to generate the textual summary but also needs to choose the video cover.

\textbf{Textual summary generation.}
For the first task, we use the final state of the input text representation $h^x_{T_d}$ as the initial state $d_0$ of the RNN decoder, and the $t$-th generation procedure is:
\begin{align}
    d_t={\rm \text{LSTM}_{dec}}(d_{t-1}, [e(y_{t-1}); h^{c}_{t-1}]),
\end{align}
where $d_t$ is the hidden state of the $t$-th decoding step and $h^{c}_{t-1}$ is the context vector calculated by the standard attention mechanism~\cite{bahdanau2014neural}, and is introduced below.

To take advantage of the article representation $h^x_t$ and the video-aware article representation $\hat{h}^x_t$, we apply an ``editing gate'' $\gamma_e$ to decide how much information of each side should be focused on:
\begin{align}
    \gamma_e &=\sigma\left( F_d(d_t) \right),\\
    g_i &=\gamma_e h^x_i + (1-\gamma_e) \hat{h}^x_i.
\end{align}
Then the context vector $h^c_{t-1}$ is calculated as:
\begin{align}
    \delta_{it}&=\frac{{\rm exp}(F_a(g_i, d_t))}{\sum\nolimits_j {\rm exp}(F_a(g_j, d_t))}.\\
    h_t^{c}&=\sum\nolimits_i \delta_{it}g_i
    \label{contextvector},
\end{align}
Finally, the context vector $h^c_t$ is concatenated with the decoder state $d_t$ and fed into a linear layer to obtain the generated word distribution $P_v$:
\begin{align}
    d_t^o&= \sigma\left( F_p([d_t;h^c_t]) \right),\\
    P_v&={\rm softmax} \left(F_o(d_t^o) \right).
\end{align}

Following \citet{See2017GetTT}, we also equip our model with pointer network to handle the out-of-vocabulary problem.
The loss of textual summary generation is the negative log likelihood of the target word $y_t$:
\begin{align}
    \mathcal{L}_{seq} = -\sum\nolimits_{t=1}^{T_y} \log P_v(y_t).
\end{align}

\textbf{Cover frame selector.}
The cover frame is chosen based on hierarchical video representations, \ie the original frame representation $M_j^i$ and the conditional segment representation $S^{c}_i$ with the article-aware segment representation $\hat{S}^{c}_i$:
\begin{align}
    p_j^i &=\gamma_f^1 S^{c}_i + \gamma_f^2 \hat{S}^{c}_i + (1-\gamma_f^1 -\gamma_f^2) M_j^i,\\
    y^{c}_{i,j} &= \sigma\left( F_c(p_j^i) \right),
\end{align}
where $y^{c}_{i,j}$ is the matching score of the candidate frames.
The fusion gates $\gamma_f^1$ and $\gamma_f^2$ here are determined by the last text encoder hidden state $h_{T_d}^x$:
\begin{align}
    \gamma_f^1 &=\sigma\left( F_{m}(h_{T_d}^x) \right),\\
    \gamma_f^2 &=\sigma\left( F_{n}(h_{T_d}^x) \right).
\end{align}

We use pairwise hinge loss to measure the selection accuracy:
\begin{align}
    \mathcal{L}_{pic} = \sum\nolimits^N \text{max} \left( 0, y^{c}_\text{negative}-y^{c}_\text{positive}+\text{margin} \right),
\end{align}
where $y^{c}_{negative}$ and $y^{c}_{positive}$ corresponds to the matching score of the negative samples and the ground truth frame, respectively.
The margin in the $\mathcal{L}_{pic}$ is the rescale margin in hinge loss.

The overall loss for the model is:
\begin{align}
    \mathcal{L} = \mathcal{L}_{seq} + \mathcal{L}_{pic}.
\end{align}

\section{Experimental Setup}
    
\subsection{Dataset}
\label{dataset}
To our best knowledge, there is no existing large-scale dataset for VMSMO task.
Hence, we collect the first large-scale dataset for VMSMO task from Weibo, the largest social network website in China.
Most of China's mainstream media have Weibo accounts, and they publish the latest news in their accounts with lively videos and articles.
Correspondingly, each sample of our data contains an article with a textual summary and a video with a cover picture.
The average video duration is one minute and the frame rate of video is 25 fps.
For the text part, the average length of article is 96.84 words and the average length of textual summary is 11.19 words.
Overall, there are 184,920 samples in the dataset, which is split into a training set of 180,000 samples, a validation set of 2,460 samples, and a test set of 2,460 samples.

\subsection{Comparisons}
\label{comparison}
We compare our proposed method against summarization baselines and VQA baselines.

\noindent \textit{Traditional Textual Summarization baselines:}

\noindent \textbf{Lead}: selects the first sentence of article as the textual summary \cite{nallapati2017summarunner}.

\noindent \textbf{TexkRank}: a graph-based extractive summarizer which adds sentences as nodes and uses edges to weight similarity \cite{Mihalcea2004TextRankBO}.

\noindent \textbf{PG}: a sequence-to-sequence framework combined with attention mechanism and pointer network \cite{See2017GetTT}.

\noindent \textbf{Unified}: a model which combines the strength of extractive and abstractive summarization \cite{hsu2018unified}.

\noindent \textbf{GPG}: \citet{shen2019improving} proposed to generate textual summary by ``editing'' pointed tokens instead of hard copying.

\noindent \textit{Multimodal baselines:}

\noindent \textbf{How2}: a model proposed to generate textual summary with video information \cite{palaskar2019multimodal}.

\noindent \textbf{Synergistic}: a  image-question-answer synergistic network to value the role of the answer for precise visual dialog\cite{guo2019image}.

\noindent \textbf{PSAC}: a model adding the positional self-attention with co-attention on VQA task \cite{li2019beyond}.

\noindent \textbf{MSMO}: the first model on multi-output task, which paid attention to text and images during generating textual summary and used coverage to help select picture \cite{zhu2018msmo}.

\noindent \textbf{MOF}: the model based on MSMO which added consideration of image accuracy as another loss \cite{zhu3multimodal}.

\subsection{Evaluation Metrics}
The quality of generated textual summary is evaluated by standard full-length Rouge F1~\cite{lin2004rouge} following previous works \cite{See2017GetTT, chen2018iterative}. 
R-1, R-2, and R-L refer to unigram, bigrams, and the longest common subsequence respectively.
The quality of chosen cover frame is evaluated by mean average precision (MAP) \cite{zhou2018multi-turn} and recall at position ($R_n@k$) \cite{tao2019multi-representation}. 
$R_n@k$ measures if the positive sample is ranked in the top $k$ positions of $n$ candidates.

\begin{table}[t]
    \centering
    \small
    \begin{tabular}{@{}lccc@{}}
        \toprule
        & R-1 & R-2 & R-L\\
        \midrule
        \emph{extractive summarization}\\
        Lead & 16.2 & 5.3 & 13.9 \\
        TextRank & 13.7 & 4.0 & 12.5 \\
        \midrule
        \emph{abstractive summarization}\\
        PG \cite{See2017GetTT} & 19.4 & 6.8 & 17.4\\
        Unified \cite{hsu2018unified} & 23.0 & 6.0 & 20.9\\
        GPG \cite{shen2019improving}  & 20.1 & 4.5 & 17.3\\
        \midrule
        \emph{our models}\\
        \textbf{DIMS} & \textbf{25.1} & \textbf{9.6} & \textbf{23.2} \\
        \bottomrule
    \end{tabular}
    \caption{Rouge scores comparison with traditional textual summarization baselines.}
    \label{tab:sum_baslines}
\end{table}

\begin{table*}[t]
    \centering
    \small
    \begin{tabular}{@{}lccccccc@{}}
        \toprule
        & R-1 & R-2 & R-L & MAP & $R_{10}@1$ & $R_{10}@2$ & $R_{10}@5$\\
        \midrule
        \emph{video-based summarization}\\
        How2 \cite{palaskar2019multimodal} & 21.7 & 6.1 & 19.0 & - & - & - & - \\
        \midrule
        \emph{Visual Q\&A methods}\\
        Synergistic \cite{guo2019image} & - & - & - & 0.588 & 0.444 & 0.557 & 0.759 \\
        PSAC \cite{li2019beyond} & - & - & - & 0.524 & 0.363 & 0.481 & 0.730 \\
        \midrule
        \emph{multimodal summarization with multimodal output}\\
        MSMO \cite{zhu2018msmo} & 20.1 & 4.6 & 17.3 & 0.554 & 0.361 & 0.551 & 0.820 \\
        MOF \cite{zhu3multimodal} & 21.3 & 5.7 & 17.9 & 0.615 & 0.455 & 0.615 & 0.817 \\
        \midrule
        \emph{our models}\\
        \textbf{DIMS} & \textbf{25.1} & \textbf{9.6} & \textbf{23.2} & \textbf{0.654} & \textbf{0.524} & \textbf{0.634} & \textbf{0.824} \\
        DIMS-textual summary & 22.0 & 6.3 & 19.2 & - & - & - & - \\
        DIMS-cover frame & - & - & - & 0.611 & 0.449 & 0.610 & 0.823\\
        \midrule
        \emph{ablation study}\\
        DIMS-G & 23.7 & 7.4 & 21.7 & 0.624 & 0.471 & 0.619 & 0.819\\
        DIMS-S & 24.4 & 8.9 & 22.5 & 0.404 & 0.204 & 0.364 & 0.634\\
        \bottomrule
    \end{tabular}
    \caption{Rouge and Accuracy scores comparison with multimodal baselines.}
    \label{tab:comp_baslines}
\end{table*}

\subsection{Implementation Details}
We implement our experiments in Tensorflow~\cite{abadi2016tensorflow} on an NVIDIA GTX 1080 Ti GPU. 
The code for our model is available online\footnote{https://github.com/yingtaomj/VMSMO}.
For all models, we set the word embedding dimension and the hidden dimension to 128.
The encoding step is set to 100, while the minimum decoding step is 10 and the maximum step is 30.
For video preprocessing, we extract one of every 120 frames to obtain 10 frames as cover candidates.
All candidates are resized to 128x64.
We regard the frame that has the maximum cosine similarity with the ground truth cover as the positive sample, and others as negative samples.
Note that the average cosine similarity of positive samples is 0.90, which is a high score, demonstrating the high quality of the constructed candidates.
In the conditional self-attention mechanism, the stacked layer number is set to 2.
For hierarchical encoding, each segment contains 5 frames.
Experiments are performed with a batch size of 16.
All the parameters in our model are initialized by Gaussian distribution.
During training, we use Adagrad optimizer as our optimizing algorithm and we also apply gradient clipping with a range of $[-2,2]$.
The vocabulary size is limited to 50k.
For testing, we use beam search with beam size 4 and we decode until an end-of-sequence token is reached.
We select the 5 best checkpoints based on performance on the validation set and report averaged results on the test set.

\begin{table}[t]
	\centering
	\small
	\begin{tabular}{@{}lcc@{}}
		\toprule
		& QA(\%) & Rating \\
		\midrule
		How2 & 46.2  &  -0.24\\
		MOF & 51.3 & -0.14 \\
		Unified & 53.8  &  0.00\\
		\textbf{DIMS} & \textbf{66.7} &\textbf{0.38} \\
		\bottomrule
	\end{tabular}
	\caption{System scores based on questions answered by human and summary quality rating.
	}
	\label{tab:comp_human_baslines}
\end{table}

\section{Experimental Result}
\subsection{Overall Performance}
We first examine \textit{whether our DIMS outperforms other baselines} as listed in Table~\ref{tab:sum_baslines} and Table~\ref{tab:comp_baslines}.
Firstly, abstractive models outperform all extractive methods, demonstrating that our proposed dataset is suitable for abstractive summarization.
Secondly, the video-enhanced models outperform traditional textural summarization models, indicating that video information helps generate summary.
Finally, our model outperforms \texttt{MOF} by 17.8\%, 68.4\%, 29.6\%, in terms of Rouge-1, Rouge-2, Rouge-L, and 6.3\%, 15.2\% in MAP and $R@1$
respectively, which proves the superiority of our model. All our Rouge scores have a 95\% confidence interval of at most $\pm$0.55 as reported by the official Rouge script.

In addition to automatic evaluation, system performance was also evaluated on the generated textual summary by human judgments on 70 randomly selected cases similar to \citet{liu2019hierarchical}.
Our first evaluation study quantified the degree to which summarization models retain key information from the articles following a question-answering (QA) paradigm \cite{Narayan2018RankingSF}.
A set of questions was created based on the gold summary. Then we examined whether participants were able to answer these questions by reading system summaries alone. We created 183 questions in total varying from two to three questions per gold summary. Correct answers were marked with 1 and 0 otherwise. The average of all question scores is set to the system score.

Our second evaluation estimated the overall quality of the textual summaries by asking participants to rank them according to its \textit{Informativeness} (does the summary convey important contents about the topic in question?), \textit{Coherence} (is the summary fluent and grammatical?), and \textit{Succinctness} (does the summary avoid repetition?).
Participants were presented with the gold summary and summaries generated from several systems better on autometrics and were asked to decide which was the best and the worst.
The rating of each system was calculated as the percentage of times it was chosen as best minus the times it was selected as worst, ranging from -1 (worst) to 1 (best).

Both evaluations were conducted by three native-speaker annotators.
Participants evaluated summaries produced by \texttt{Unified}, \texttt{How2}, \texttt{MOF} and our \texttt{DIMS}, all of which achieved high perfromance in automatic evaluations.
As shown in Table \ref{tab:comp_human_baslines}, on both evaluations, participants overwhelmingly prefer our model.
All pairwise comparisons among systems are statistically significant using the paired student t-test for significance at $\alpha$ = 0.01.

\begin{figure*}
    \centering
    \includegraphics[scale=0.55]{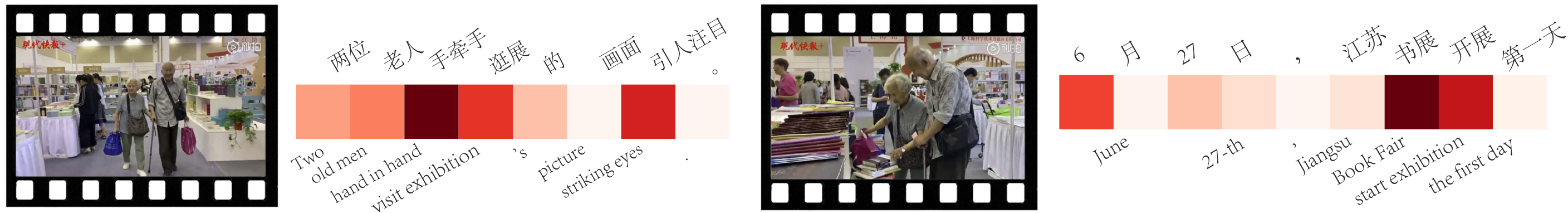}
    \caption{
        Visualizations of global-attention matrix between the news article and two frames in the same video.
    }
    \label{fig:global_visual}
\end{figure*} 

\subsection{Ablation Study}

Next, we conduct ablation tests to assess the importance of the conditional self-attention mechanism (-S), as well as the global-attention (-G) in Table~\ref{tab:comp_baslines}.
All ablation models perform worse than \texttt{DIMS} in terms of all metrics, which demonstrates the preeminence of \texttt{DIMS}.
Specifically, the global-attention module contributes mostly to the textual summary generation, while the conditional self-attention module is more important for choosing cover frame.


\subsection{Analysis of Multi-task learning}
Our model aims to generate textural summary and choose cover frame at the same time, which can be regarded as a multi-task.
Hence, in this section, we examine whether these two tasks can complement each other.
We separate our model into two single-task architecture, named as \texttt{DIMS-textual summary} and \texttt{DIMS-cover frame}, which generates textural summary and chooses video cover frame, respectively.
The result is shown in Table~\ref{tab:comp_baslines}.
It can be seen that the multi-task \texttt{DIMS} outperforms single-task \texttt{DIMS-textual summary} and \texttt{DIMS-cover frame}, improving the performance of summarization by 20.8\% in terms of ROUGE-L score, and increasing the accuracy of cover selection by 7.0\% on MAP.

\subsection{Visualization of dual interaction module}
To study the multimodal interaction module, we visualize the global-attention matrix $E^{t}_{i}$ in Equation~\ref{global} on one randomly sampled case, as shown in Figure \ref{fig:global_visual}. In this case, we show the attention on article words of two representative images in the video.
The darker the color is, the higher the attention weight is.
It can be seen that for the left figure, the word \emph{hand in hand} has a higher weight than \emph{picture}, while for the right figure, the word \emph{Book Fair} has the highest weight.
This corresponds to the fact that the main body of the left frame is two old men, and the right frame is about reading books.

We show a case study in Table \ref{tab:case_study}, which includes the input article and the generated summary by different models.
We also show the question-answering pair in human evaluation and the chosen cover.
The result shows that the summary generated by our model is both fluent and accurate, and the cover frame chosen is also similar to the ground truth frame.

\begin{CJK*}{UTF8}{gkai}
\begin{table}[t]
    \centering
    \small
     \begin{tabular}{ll}
      \toprule
      \multicolumn{2}{p{7cm}}{    \emph{\textbf{Article:}}
      On August 26, in Shanxi Ankang, a 12-year-old junior girl Yu Taoxin goose-stepped like parade during the military training in the new semester, and won thousands of praises. 
      Yu Taoxin said that her father was a veteran, and she worked hard in military training because of the influence of her father. 
      Her father told her that military training should be strict as in the army.
        8月26日，陕西安康，12岁的初一女生余陶鑫，在新学期军训期间，她踢出阅兵式般的标准步伐，获千万点赞。余陶鑫说，爸爸是名退伍军人，军训刻苦是因为受到爸爸影响，爸爸曾告诉她，军训时就应和在部队里一样，严格要求自己。
      } \\ \hline 
      \multicolumn{2}{p{7cm}}{
        \emph{\textbf{Reference summary:}}
        A 12-year-old girl goose-stepped like parade during the military training, ``My father is a veteran.''
        12岁女孩军训走出阅兵式步伐，“爸爸是退伍军人”
        } \\ \hline
        \multicolumn{2}{p{7cm}}{
         \textbf{QA:} \emph{What happened on the 12-year-old girl?}
         [\emph{She goose-stepped like parade.}]
         这个12岁女孩做了什么？[她走出阅兵式步伐。]
         }\\
         \multicolumn{2}{p{7cm}}{
         \emph{Why did she do this?}
         [\emph{She was influenced by her father}]
         她为什么这样做？[她受到爸爸的影响。]
         }\\ \hline
      \multicolumn{2}{p{7cm}}{
        \emph{\textbf{Unfied:}}
        12-year-old gril Yu Taoxin goose-stepped during military training.
        12岁女生余陶鑫军训期间阅兵式般的标准步伐}\\ \hline
      \multicolumn{2}{p{7cm}}{
        \emph{\textbf{How2:}}
        12-year-old girls were organized military training, and veteran mother parade.
        12岁女生组团军训，退伍军人妈妈阅兵式}\\ \hline
      \multicolumn{2}{p{7cm}}{
        \emph{\textbf{MOF:}}
        A 12-year-old junior citizen [unk]: father gave a kicked like.
        1名12岁初一市民 [unk]：爸爸踢式点赞}\\ \hline
        \multicolumn{2}{p{7cm}}{
        \emph{\textbf{DIMS:}}
        A 12-year-old junior girl goose-stepped like parade: My father is a veteran, and military training should be strict as in the army.
        12岁初一女生踢出阅兵式：爸爸是名退伍军人，军训时就应和在部队一样}\\ \hline
        \multicolumn{2}{p{7cm}}{
        \begin{minipage}{0.1\textwidth}
            \includegraphics[scale=0.5]{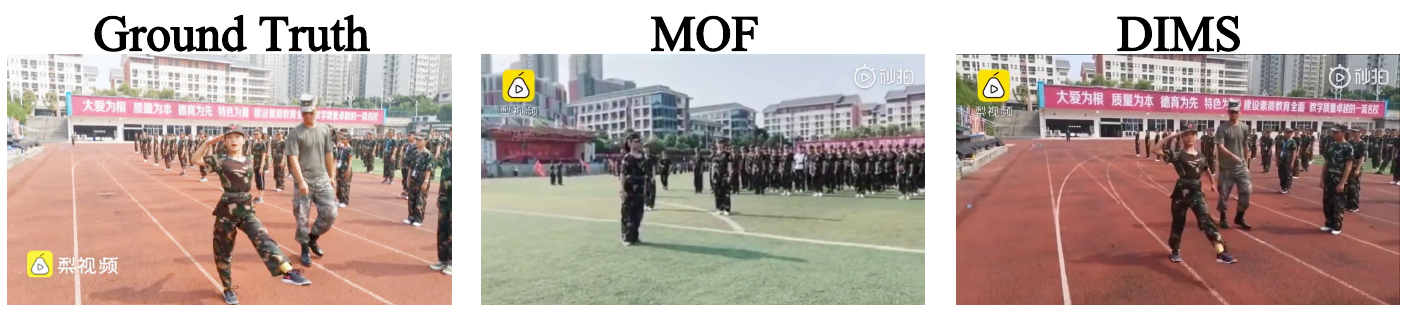}
        \end{minipage}} \\
      \bottomrule
    \end{tabular}

    \caption{Examples of the generated summary by baselines and DIMS.
    }
    \label{tab:case_study}
\end{table}
\end{CJK*}
    
\section{Conclusion}
In this paper, we propose the task of Video-based Multimodal Summarization with Multimodal Output (VMSMO) which chooses a proper video cover and generates an appropriate textual summary for a video-attached article.
We propose a model named Dual-Interaction-based Multimodal Summarizer (DIMS) including a local conditional self-attention mechanism and a global-attention mechanism to jointly model and summarize multimodal input.
Our model achieves state-of-the-art results in terms of autometrics and outperforms human evaluations by a large margin.
In near future, we aim to incorporate the video script information in the multimodal summarization process.

\section*{Acknowledgments}

We would like to thank the anonymous reviewers for their constructive comments. 
This work was supported by the National Key Research and Development Program of China (No.2020AAA0105200), and the National Science Foundation of China (NSFC No.61876196, No.61672058).
Rui Yan is partially supported as a Young Fellow of
Beijing Institute of Artificial Intelligence (BAAI).


\end{document}